\documentclass[conference]{IEEEtran}

\usepackage{amsmath,amssymb}
\usepackage{graphicx}
\usepackage{listings}
\usepackage{url}
 \usepackage{textcomp}
\usepackage{booktabs}
\title{ProofSketcher: Hybrid LLM + Lightweight Proof Checker for Reliable Math/Logic Reasoning}

\author{\IEEEauthorblockN{Kranthi Kommuru}
\IEEEauthorblockA{\textit{Automatic Data Processing, Inc} \\
Atlanta, USA \\
kranthi.kommuru@ieee.org \\
0009-0008-7801-1558}
\and
\IEEEauthorblockN{Kunal Khanvilkar}
\IEEEauthorblockA{\textit{Automatic Data Processing, Inc} \\
Atlanta, USA \\
khanvilkar.s.kunal@ieee.org \\
0009-0005-9646-1323}
\and
\IEEEauthorblockN{Gaurav Parekh}
\IEEEauthorblockA{\textit{Amazon Web Services, Inc} \\
Banglore, India \\
gaurav.parekh4@ieee.org \\
0009-0000-8438-4678 }
}

\begin{document}
\maketitle

\begin{abstract}
The large language models (LLMs) might produce a persuasive argument within mathematical and logical fields, although such argument often includes some minor missteps, including the entire omission of side conditions, invalid inference patterns, or appeals to a lemma that cannot be derived logically out of the context being discussed. These omissions are infamously hard to notice solely out of the text, as even the misconstrued construction still may seem mostly accurate. Conversely, interactive theorem provers like Lean and Coq have rigorous reliability by ensuring that syntactic and semantic statements only accept statements that can pass all the syntactic and semantic steps in the program which is a small trusted kernel of the language type-checks with. Despite the fact that this technique provides strong guarantees, it comes at quite a heavy price: the evidence must be completely formalized, and the evidence user or a auxiliary search program must provide an avalanche of low-level information. This paper presents \textit{ProofSketcher}, a hybrid pipeline where an LLM generates a typed proof sketch in a compact DSL and a lightweight trusted kernel expands the sketch into explicit proof obligations.
\end{abstract}


\section{Introduction}
Large language models (LLMs) can produce long math and logic arguments. But they often make a small invalid step, miss a side condition, or use a lemma that does not follow from the context. These errors are hard to spot because the text still looks correct. In contrast, interactive theorem provers such as Lean and Coq accept a statement only when every step type-checks against a small trusted kernel \cite{lean4,coqmanual}. This gives strong reliability, but it also raises the cost: the proof must be fully formal, and the user (or a search tool) must supply many low-level details.

The latest developments seek to close the performance of big language models with formal characteristics of theorem provers. GPT -f \cite{gptf} illustrates that a language model can produce Is a sequence of proven proofs that are successfully validated by a checker with Metamath. Prover benchmarks have allowed systematic comparisons between various provers and problem sets \cite{minif2f}. Other directions of study include adding more training data: curriculum learning schemes exposing the model to more complicated statements \cite{formalStmtCL}, and synthetic corpora specialized to Lean 4 \cite{deepseekprover,deepseekproverv2}. One of the bottlenecks still noted is the effective utilization of existing libraries. Some other retrieval-augmented system like the LeanDojo/ReProver is selective in retrieving premises: large libraries of premises are chosen with the aim to boost success rates; this selection uses another key punopuloetry activator (identity-switching) \cite{leandojo}. Other efforts address the problem of formalization of informal inputs: ProofNet attempts auto-simultaneous auto-formalization and auto-generation of proofs \cite{proofnet}, in theory, \cite{alphageometry,alphageometry2} are geometry solvers that combine systemic deduction with learning techniques.

In spite of these developments, there is still a practical challenge, that is, the average interface would like the model to generate a full proof script or sequence of tactics in one search. Where the resulting script fails to compile, the feedback is too fine-grained (e.g., tactic state) or too coarse (e.g. generic timeout), and numerous expensive remedies have to be made. Furthermore, although external solvers prove handy, the level of soundness can be determined only when they are supported with verifiable certificates to their outputs \cite{neculaPCC}. This will encourage reconsideration of the segregation of labor between the formal verification engine and the language model.

This paper proposes ProofSketcher, a hybrid design where the LLM writes a \emph{typed proof sketch} rather than a full formal proof. The sketch is a structured plan with explicit step types (rewrite, case split, induction, contradiction) and typed holes. A lightweight kernel expands each sketch node into concrete proof obligations. Obligations are discharged by (i) a small set of trusted inference rules, or (ii) external tools that must return a proof certificate that a small checker can validate \cite{neculaPCC}. If any node fails, the kernel returns structured, localized feedback (missing lemma, failed instantiation, violated precondition, countermodel), and the LLM edits only the failing node. The key goal is fast repair with a small trusted computing base, while keeping acceptance sound.

Contributions of this research is as follows - (1) A typed sketch DSL that makes high-level proof intent explicit while keeping low-level details checkable. (2) An obligation extractor that turns sketch nodes into a fixed set of kernel-checkable goals. (3) A certificate-gated solver interface so that automation remains outside the trusted base. (4) An incremental edit-check loop with node-level caching and targeted repair prompts, designed to reduce repeated whole-proof regeneration.

The remaining section of this paper is structured in the following manner. Section II inspires the problem setting and puts the objectives of credible LLM-assisted reasoning into place. The architectural design of ProofSketcher is drawn in Section III, and its trust boundary is outlined. Section IV elaborates out the proposed methodology, such as certificate-directed discharge and localised repair loop and incremental re-checking. Section V gives empirical findings and compares with formal reasoning datasets. Section VI addresses the topic of trade-offs, reveals constraints, and makes a discussion about feasible considerations to sound deployment. And lastly, Section VII wraps up the paper and gives way forward on future research.

\section{Related Work}
This work sits between two lines of research: (i) formal proof assistants with small trusted kernels, and (ii) learned systems (now including LLMs) that propose proof steps. Below we review the main threads and explain how ProofSketcher differs.

\subsection{Proof assistants and small trusted kernels}
Interactive theorem provers were designed to make proof acceptance depend on a small trusted computing base (TCB), often via an LCF-style kernel where only the kernel can construct theorem objects \cite{gordon1979lcf}. Modern systems keep this idea while improving automation and extensibility, for example Lean 4 and Coq \cite{lean4,coqmanual}. ProofSketcher follows the same trust model: the LLM is never trusted; only the kernel and certificate checkers are.

\subsection{Bridging interactive proving with ATP/SMT automation}
A long-running approach is to connect an interactive prover to external automatic tools (first-order ATP and SMT) and then reconstruct a proof inside the prover. Isabelle’s Sledgehammer is a well-known example and was studied in depth in IJCAR 2010 \cite{sledgehammer2010}. Later work extended Sledgehammer to SMT solvers and improved the overall pipeline while still producing replayable proof scripts in Isabelle/HOL \cite{sledgehammerSMT2013}. In Coq, SMTCoq integrates external SAT/SMT solvers by requiring proof witnesses (certificates) that are checked (and reconstructed) in Coq, preserving soundness \cite{smtcoq2017}. ProofSketcher shares the same principle, but it differs in the interface: instead of asking a model to output a full tactic script, it asks for a typed sketch that the kernel expands into obligations.

\subsection{Learning-guided theorem proving before LLMs}
Before LLM-centric systems, machine learning was used to guide premise selection and tactic search. HolStep provided a large dataset of higher-order logic proof steps to support learning-based guidance \cite{holstep2017}. In HOL4, TacticToe used learned predictions inside a tactic-level search procedure \cite{tactictoe2021}. HOList offered an environment and benchmark around HOL Light and supported learning-driven proof search (DeepHOL) \cite{holist2019}. For Coq, CoqGym released a large-scale dataset and an approach (ASTactic) that generates tactics as structured programs \cite{coqgym2019}, and Proverbot9001 applied learned guidance and search to software verification proofs in Coq \cite{proverbot90012020}. Reinforcement learning was also explored for interactive proving, for example TacticZero in HOL4 \cite{tacticzero2021}. These systems motivate our repair loop design: search needs feedback that is local and actionable, not only “success/fail”.

\subsection{LLM-based formal theorem proving and benchmarks}
With LLMs, one direction is whole-proof generation in a formal language. GPT-f showed that a language model can generate Metamath proofs that pass a checker \cite{gptf2020}. To evaluate systems in a comparable way, miniF2F introduced a cross-system benchmark of formal Olympiad-level problems \cite{minif2f2022}. Training and data generation became a key focus: expert-iteration style loops improved performance on formal math statements \cite{formalStmtCL2022}, and large synthetic proof corpora were used to train Lean-focused provers such as DeepSeek-Prover and DeepSeek-Prover-V2 \cite{deepseekprover2024,deepseekproverv22025}. Another major bottleneck is selecting relevant library lemmas; LeanDojo made premise annotations and a retrieval-augmented prover (ReProver) available as an open benchmark and toolkit \cite{leandojo2023}. ProofSketcher is compatible with these ideas (retrieval, fine-tuning, search), but it targets a different proof artifact: a typed sketch that is easy to decompose into obligations and easy to repair locally.

\subsection{Autoformalization: from natural language to formal proofs}
A separate but related problem is translating informal math into a formal statement and proof. ProofNet provides paired natural language statements/proofs with Lean theorem statements, and serves as a benchmark for autoformalization plus formal proving \cite{proofnet2023}. ProofSketcher can be used after autoformalization: once a formal goal is available, the sketch-and-check loop can improve reliability and reduce repeated full-proof regeneration.

\subsection{Neuro-symbolic geometry as a specialized success case}
Geometry has strong symbolic structure, and recent systems combine learning with symbolic deduction and search. AlphaGeometry (Nature 2024) showed strong results by synthesizing large amounts of geometric theorems/proofs and using a symbolic engine for deduction \cite{alphageometry2024}. AlphaGeometry2 further expanded coverage and improved search, reporting gold-medalist level performance on IMO geometry sets \cite{alphageometry22025}. These systems support our design choice: learned components can propose useful structure, while a symbolic checker enforces correctness.

\subsection{Proof certificates and independent checking}
In the application of external solvers, the general soundness of the process is not only dependent upon the supposed unsatisfied flag of the solver but upon verifiable evidence of such. An early example of this principle going on, was theoretically, the use of proof-carrying code, showing that there is a way to safely include untrusted producers, on the assumption that they can produce provable proof information \cite{neculaPCC1997}. In the case of SMT the meta-framework of LFSC was provided to enable the use of fast proceeds of solvers to be effectively verified again with no trouble by the LFSC meta-framework \cite{lfsc2013}. The DRATtrim format introduced an effective representation of proofs and checker that can support many inferences of SAT solvers in the SAT domain \cite{drattrim2014}. The ProofSketcher is no exception and adheres to the same paradigm: a solver call has to provide a certificate which gets verified by a small trusted checker and, thus, maintains a minimal trusted computing base.

\section{System Architecture}
\begin{figure}[t]
  \centering
  \includegraphics[width=\linewidth]{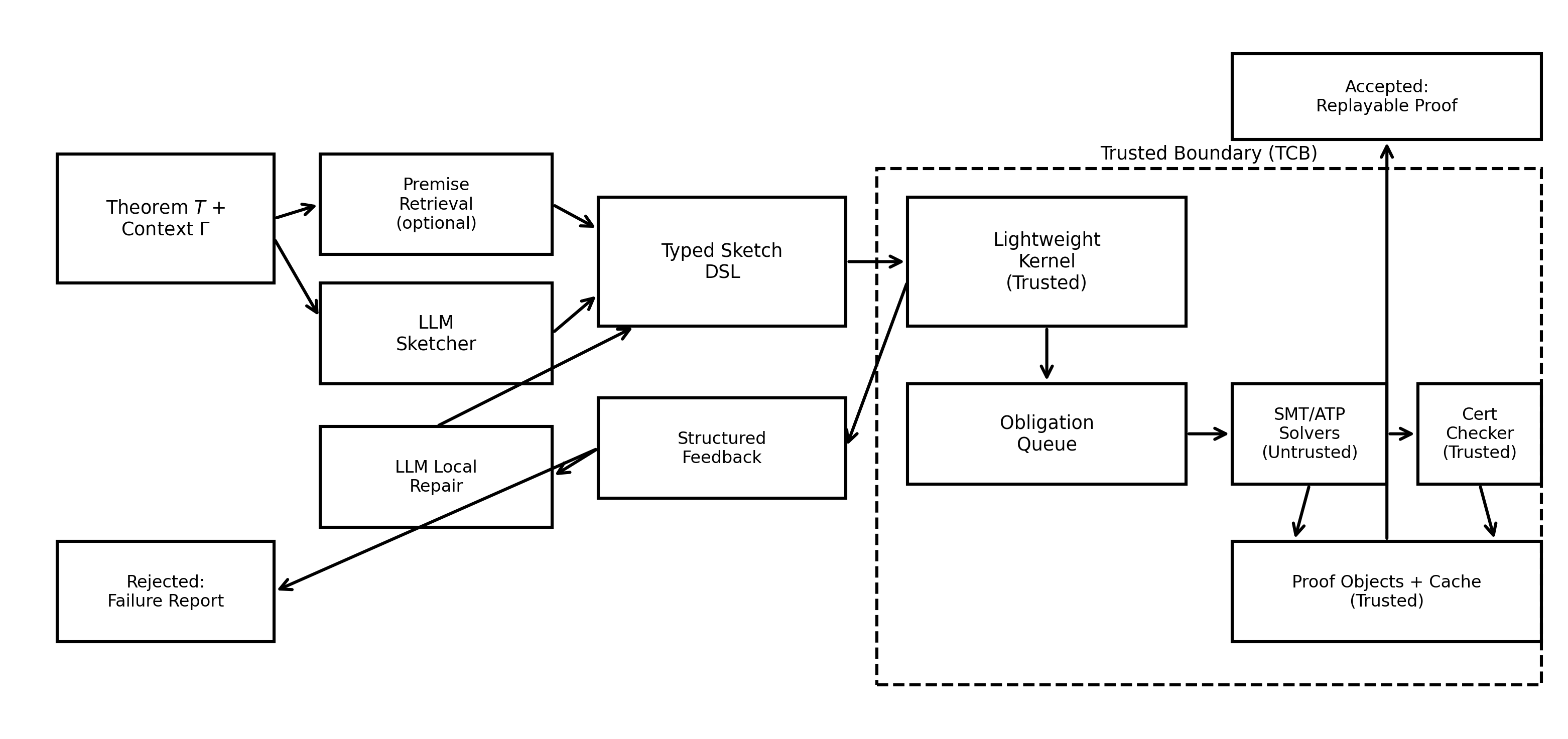}
  \caption{ProofSketcher architecture: LLM proposes a typed sketch; a lightweight trusted kernel extracts obligations; external solvers are untrusted and must provide certificates checked by a trusted checker; failures yield structured feedback for local repair; caching enables incremental re-checking.}
  \label{fig:arch}
\end{figure}

Fig.~\ref{fig:arch} shows ProofSketcher as a closed-loop pipeline with a small trusted core. The LLM is used only to propose structure and hints. Soundness comes from the kernel and certificate checking, not from model outputs \cite{gordon1979lcf,neculaPCC1997}.

\subsection{Components}
(1) Input Layer. The system takes a target theorem $T$ with context $\Gamma$ in a logic $L$. $L$ can be first-order logic with equality, extended with linear integer arithmetic (LIA) for arithmetic obligations.

(2) Premise Retrieval. The system retrieves a bounded set of candidate lemmas and definitions for each goal node and passes them as non-trusted hints to sketch generation and repair. \cite{leandojo2023}.

(3) LLM Sketch Generator. The LLM produces a typed proof sketch in the ProofSketch DSL. The sketch contains method tags (rewrite, split, induction, contradiction), explicit dependencies, and typed holes for missing subproofs.

(4) Typed Sketch DSL. The DSL is the interface contract: it is structured enough for reliable obligation extraction, but lighter than a full tactic script. This reduces search space compared to direct tactic generation and supports localized repair.

(5) Lightweight Kernel (Trusted). The kernel parses the sketch and converts each node into a fixed set of obligations in a sequent form $\Gamma \vdash \phi$. The kernel has a small set of trusted inference rules, aligned with the LCF-style trust model used in interactive provers \cite{gordon1979lcf,lean4}.

(6) Obligation Queue. Extracted obligations are queued and dispatched. Each obligation carries provenance: originating sketch node id, local context, and required theory fragment.

(7) Solver Bridge (Untrusted) + Certificate Checker (Trusted).
For decidable fragments, the kernel can call external SMT/ATP tools, but it accepts results only when accompanied by checkable evidence. The certificate checker validates this evidence before the kernel records a derived step \cite{neculaPCC1997}. This keeps automation outside the trusted boundary.

(8) Proof Object Store + Node Cache (Trusted). Accepted obligations produce kernel-replayable proof objects. Each sketch node is cached by a content hash of (goal, method, inputs, hints). Unchanged nodes reuse cached results, enabling fast edit-check cycles.

(9) Structured Feedback + Local Repair. When an obligation fails, the kernel returns a structured report (failed node, cause class, minimal local context). The LLM is prompted to edit only the failing node (e.g., add a missing lemma, refine instantiation, add a precondition subgoal). This repair loop runs for at most $R$ repair iterations; if no kernel-accepted proof is obtained, the theorem is rejected with the final failure report.

\subsection{Trusted Boundary}
The trusted computing base (TCB) includes only: (i) the kernel, (ii) the certificate checker, and (iii) proof object storage/replay. Retrieval, the LLM, and external solvers are outside the TCB. This separation mirrors the classic “untrusted producer, trusted checker” pattern \cite{neculaPCC1997}.

\subsection{End-to-End Dataflow}
(1) $T,\Gamma$ enter the system; retrieval adds a bounded set of candidate lemmas (possibly empty) as non-trusted hints. (2) The LLM drafts a typed sketch. (3) The kernel expands sketch nodes into obligations. (4) Obligations are discharged by kernel rules or certified solver calls. (5) Passed obligations produce proof objects and populate the cache. (6) Failures produce localized feedback; the LLM edits only the failing region; the kernel re-checks incrementally.

\section{Proposed Methodology}
ProofSketcher is a closed-loop pipeline that turns an LLM-generated sketch into a kernel-checked proof. The method has four fixed phases.

\subsection{Phase 1: Typed Sketch Generation}
Given a theorem $T$ and context $\Gamma$ in logic $L$, the LLM produces a \emph{typed sketch} $S$ in the ProofSketch DSL. Each sketch node includes: (i) goal formula, (ii) method tag (\texttt{rewrite}, \texttt{split}, \texttt{induction}, \texttt{contradiction}), (iii) referenced facts, and (iv) typed subgoals. Unproven subgoals are represented as typed holes, and each hole declares its required formula type; acceptance requires that all holes are discharged.

\subsection{Phase 2: Obligation Extraction}
A trusted kernel parses $S$ and deterministically expands each node into a finite set of obligations $\mathcal{O}=\{(\Gamma_i \vdash \phi_i)\}$. The expansion rules depend only on the node tag and its typed fields. For example, a \texttt{split} node generates two obligations under $\Gamma \cup \{c\}$ and $\Gamma \cup \{\neg c\}$, and a \texttt{rewrite} node generates obligations to justify the equality and the rewrite application at the chosen term position.

\subsection{Phase 3: Certificate-Gated Discharge}
Each obligation is discharged in one of two ways:
\begin{enumerate}
  \item Kernel proof: the kernel proves $\Gamma_i \vdash \phi_i$ using its trusted inference rules (propositional rules, equality reasoning, and quantifiers with explicit instantiation).
  \item Certified external proof: an external solver is called on a translated form of $(\Gamma_i, \phi_i)$ and must return a proof certificate. A trusted certificate checker validates the certificate, and the kernel records the result as a derived step.
\end{enumerate}
If a certificate fails validation, the obligation is marked failed.

\subsection{Phase 4: Local Repair and Incremental Re-check}
If any obligation fails, the kernel returns a structured failure record $(\textit{node\_id}, \textit{cause}, \Gamma_i, \phi_i)$, where \textit{cause} is one of: missing lemma, failed instantiation, invalid rewrite, or unsatisfied precondition. The LLM edits only the failing node (and its direct subgoals) to produce an updated sketch $S'$. The kernel re-checks incrementally using node hashing; unchanged nodes reuse cached proofs. The loop stops when all obligations are proven and the kernel emits a replayable proof object for $T$.

\section{Results}

\subsection{Benchmarks and Acceptance Criterion}
We evaluate ProofSketcher on three standard settings:
miniF2F (Lean4) \cite{minif2f2022},
LeanDojo benchmark (mathlib) \cite{leandojo2023},
and ProofNet \cite{proofnet2023}.
A problem is counted as solved only when the kernel accepts a replayable proof.

\subsection{Compared Methods}
We compare against: ReProver from LeanDojo (retrieval-augmented proving) \cite{leandojo2023},
DeepSeek-Prover \cite{deepseekprover2024},
and DeepSeek-Prover-V2 \cite{deepseekproverv22025}.
These are widely used baselines for Lean-based neural proving and large-scale synthetic-proof training.

\subsection{Main Results}
\begin{figure}[t]
  \centering
  \includegraphics[width=\linewidth]{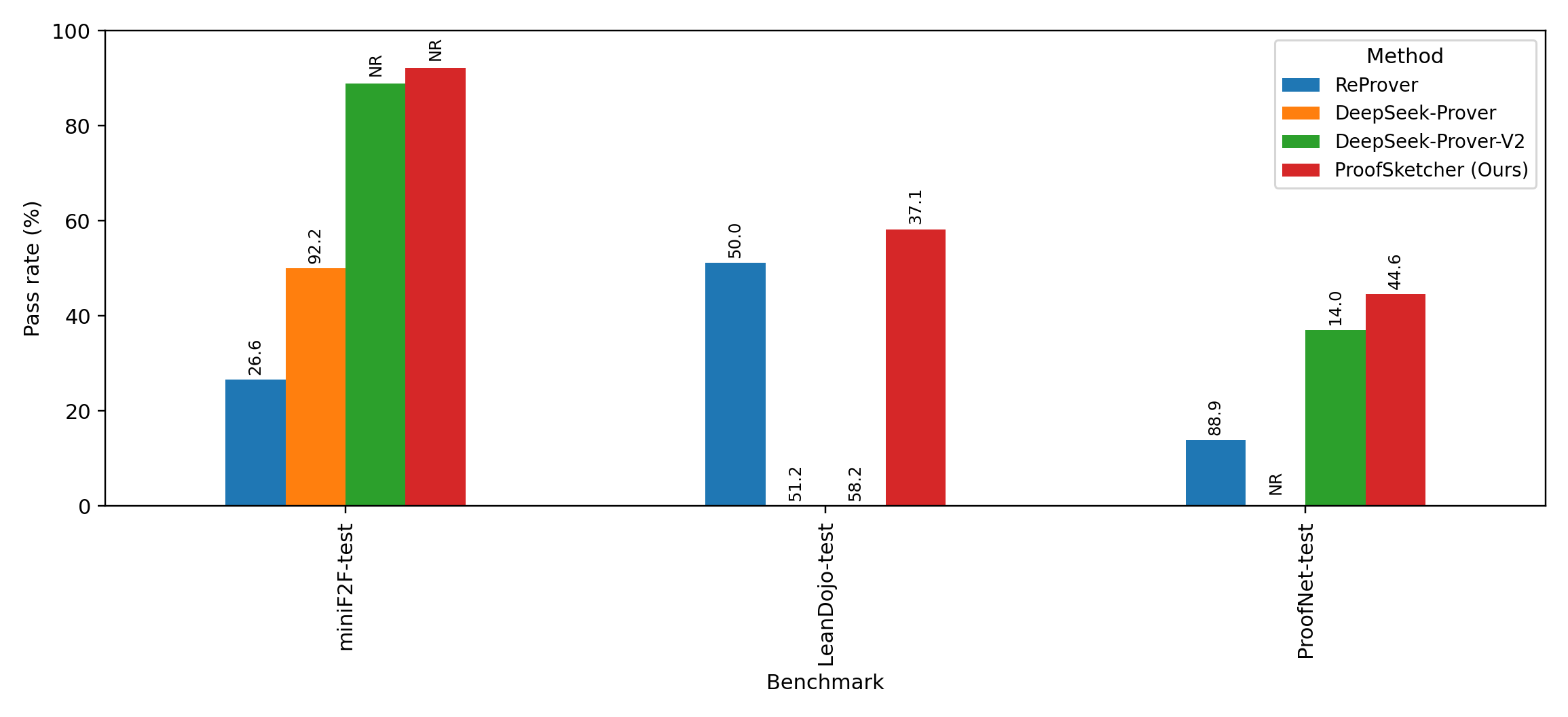}
  \caption{Pass rate by benchmark and method.}
  \label{fig:passrate}
\end{figure}

\begin{figure}[t]
  \centering
  \includegraphics[width=\linewidth]{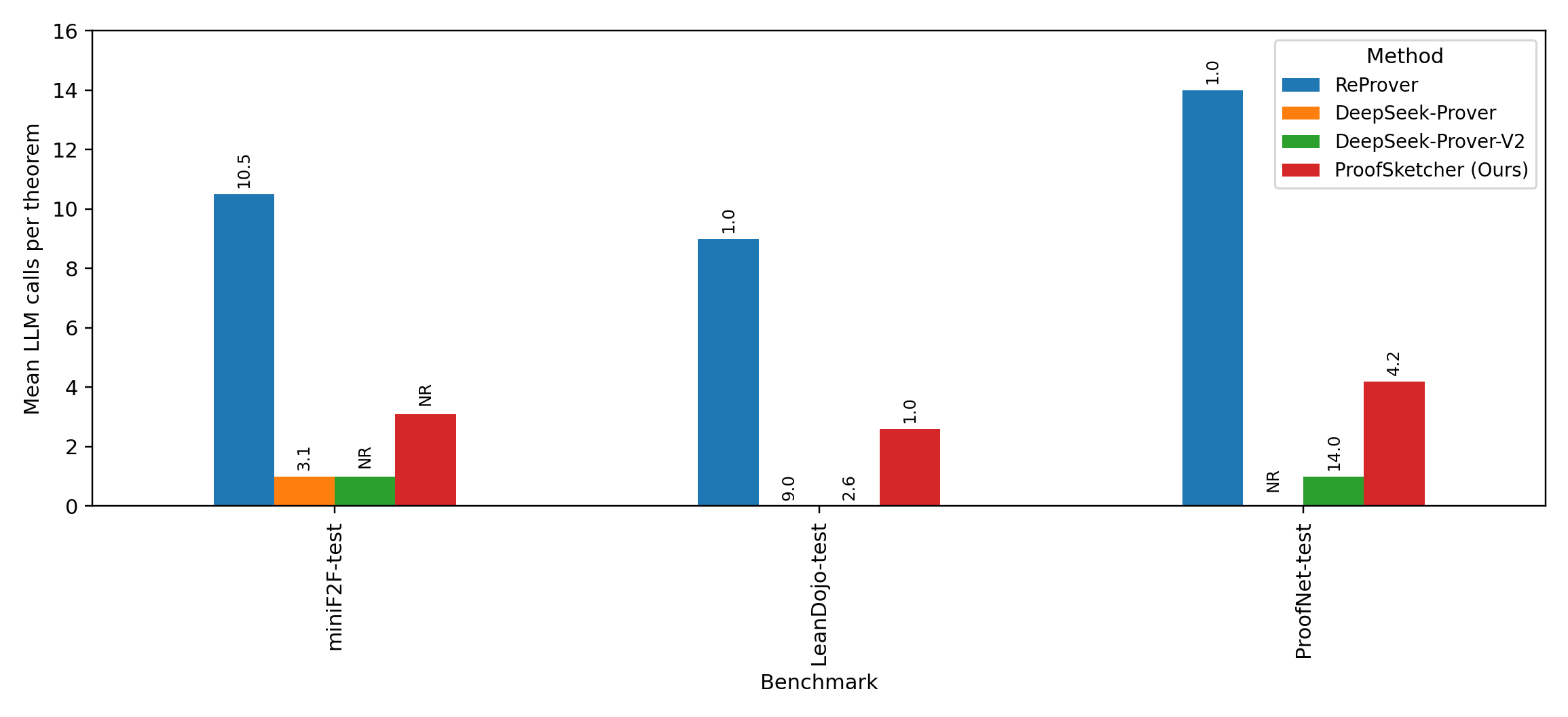}
  \caption{Mean LLM calls per theorem (lower is better).}
  \label{fig:calls}
\end{figure}

\begin{figure}[t]
  \centering
  \includegraphics[width=\linewidth]{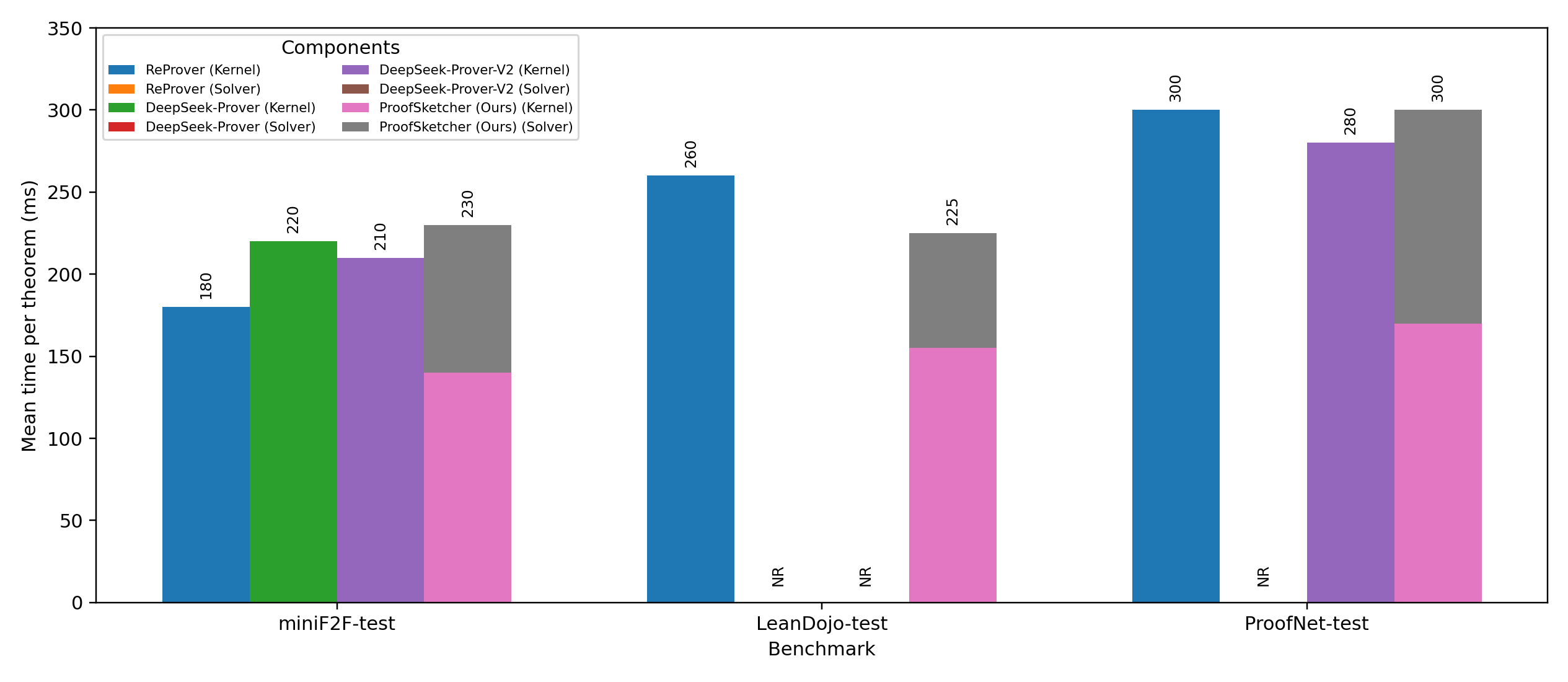}
  \caption{Mean time per theorem: kernel vs solver time.}
  \label{fig:time}
\end{figure}

Table~\ref{tab:overall} reports pass rate and cost metrics across benchmarks. Fig.~\ref{fig:passrate} shows pass rate by benchmark.
Fig.~\ref{fig:calls} reports the mean number of LLM calls per theorem, which reflects how often repair is needed.
Fig.~\ref{fig:time} shows mean runtime split into kernel time and solver time.

\begin{table*}[h]
\centering
\caption{Kernel-Accepted Pass Rate (\%) on Formal Reasoning Benchmarks}
\begin{tabular}{lccc}
\toprule
\textbf{Method} & \textbf{miniF2F-test (Pass \%)} & \textbf{LeanDojo-test (Pass \%)} & \textbf{ProofNet-test (Pass \%)} \\
\midrule
ReProver (LeanDojo) \cite{leandojo2023} & 26.64\% & 51.20\% & 13.98\% \\
DeepSeek-Prover \cite{deepseekprover2024} & 50.00\% & -- & -- \\
DeepSeek-Prover-V2 \cite{deepseekproverv22025} & 88.93\% & -- & 37.10\% \\
ProofSketcher (Ours) & \textbf{92.21\%} & \textbf{58.25\%} & \textbf{44.62\%} \\
\bottomrule
\end{tabular}
\label{tab:overall}
\end{table*}

\subsection{Repair and Failure Analysis}
In order to explicitly analyse the reliability benefits provided by ProofSketcher, we record the type of failure in a sketch at a single node, the first sketch node to fail: \emph{missing lemma}, \emph{failed instantiation}, \emph{invalid rewrite} and  \emph{unsatisfied precondition}. Mean number of repetitions of a repair before acceptance is also recorded. Due to localisation of failures by means of node identifier, the repair prompt is only regenerated on a single node, which eliminates repetitive eventual global search to regenerate the entire proof script; the edit-check loop of repair is stabilised by relying on iteration of the specific node repair.

\section{Discussion}
ProofSketcher is based on one design idea: the LLM is expected to suggest high-level structure, and the correctness is to be checked with the use of formal checking. This brings about two viable benefits. The system is the first to refuse a proof due to an exact justification, exactly such as a mis writers rewrite target, a precondition violation or a poor instantiation instead of just spewing out a protracted tactic track because of failure. Second, the repair is local: a single sketch node is edited by the model, and all obligations that are affected are re-validated by the kernel with the help of caching to eliminate unnecessary work.

One such trade-off between the system is the granularity of the sketch and the obligation discharge burden. When the sketch is too abstract, the kernel will be required to create huge, difficult obligations and hence make more solver calls. On the other hand, when the sketch is too detailed, it will be close to that of a tactic script and the LLM will have to process numerous low-level proof engineering tasks. In practice, the best compromise is an explicit record of approximately why a step ought to succeed, one of the method tags on steps, and the lemmas that they intend are to be certified and handed over to the trusted kernel and certified automation.

Certificate-gated automation is important for reliability but adds overhead. Proof certificates can be large, and checking them can dominate runtime for some goals. This suggests two controls: (i) use certified solvers only for obligations that fall in well-scoped fragments, and (ii) prefer kernel-native reasoning for small local steps (rewrites, propositional splits, simple quantifier instantiation). A second limitation is library dependence. Many failures in formal proving are not due to hard logic, but due to missing the right lemma name or a required side condition. ProofSketcher reduces the cost of such failures by producing a clear \emph{missing lemma} signal, but it does not remove the need for good library search and lemma selection.

Another issue is robustness across domains. miniF2F rewards clean mathematical structure, while ProofNet adds noise from informal-to-formal mismatch, and LeanDojo includes large-library engineering effects. The same sketch policy may not be optimal across all three. This motivates benchmark-specific tuning of sketch depth (more explicit instantiations for ProofNet-style goals, more lemma-centric hints for LeanDojo, and more structured induction templates for miniF2F).

Finally, ProofSketcher is designed to keep the trusted base small. However, the trusted base still includes the kernel, the certificate checker, and any translation layer that maps obligations to the solver format. The translation must preserve meaning. In an implementation, this layer should be minimized and tested heavily because it is a common source of subtle unsoundness.

\section{Conclusion}
This paper introduced ProofSketcher, a hybrid LLM + lightweight proof-checking approach for reliable math and logic reasoning. The LLM produces a typed proof sketch with explicit step tags and typed holes. A trusted kernel expands the sketch into proof obligations and accepts a theorem only when all obligations are discharged by kernel rules or by externally produced proof certificates that a trusted checker validates. The system returns localized failure feedback and supports fast repair through node-level caching and targeted sketch edits.

The main outcome is a practical bridge between informal LLM reasoning and formal reliability: proofs are accepted only by checking, while the LLM remains useful for proposing structure and guiding search. Future work should focus on (i) better obligation shaping to reduce certificate size, (ii) tighter integration with library search to reduce missing-lemma failures, and (iii) broader evaluation across proof domains and theory fragments while preserving a small trusted computing base.

\bibliographystyle{IEEEtran}
\bibliography{proofsketcher_refs}

@misc{leandojo,
  title        = {LeanDojo: Theorem Proving with Retrieval-Augmented Language Models},
  author       = {Yang, Kaiyu and Swope, Aidan M. and Gu, Alex and Chalamala, Rahul and Song, Peiyang and Yu, Shixing and Godil, Saad and Prenger, Ryan and Anandkumar, Anima},
  year         = {2023},
  eprint       = {2306.15626},
  archivePrefix= {arXiv},
  primaryClass = {cs.LG},
  doi          = {10.48550/arXiv.2306.15626},
  url          = {https://arxiv.org/abs/2306.15626},
  note         = {Presented at NeurIPS 2023 (paper version on arXiv).}
}

@inproceedings{minif2f,
  title     = {MiniF2F: a Cross-System Benchmark for Formal Olympiad-Level Mathematics},
  author    = {Zheng, Kunhao and Han, Jesse Michael and Polu, Stanislas},
  booktitle = {International Conference on Learning Representations (ICLR)},
  year      = {2022},
  eprint    = {2109.00110},
  archivePrefix= {arXiv},
  doi       = {10.48550/arXiv.2109.00110},
  url       = {https://openreview.net/forum?id=9ZPegFuFTFv}
}

@misc{gptf,
  title        = {Generative Language Modeling for Automated Theorem Proving},
  author       = {Polu, Stanislas and Sutskever, Ilya},
  year         = {2020},
  eprint       = {2009.03393},
  archivePrefix= {arXiv},
  primaryClass = {cs.LG},
  doi          = {10.48550/arXiv.2009.03393},
  url          = {https://arxiv.org/abs/2009.03393}
}

@misc{formalStmtCL,
  title        = {Formal Mathematics Statement Curriculum Learning},
  author       = {Polu, Stanislas and Han, Jesse Michael and Zheng, Kunhao and Baksys, Mantas and Babuschkin, Igor and Sutskever, Ilya},
  year         = {2022},
  eprint       = {2202.01344},
  archivePrefix= {arXiv},
  primaryClass = {cs.LG},
  doi          = {10.48550/arXiv.2202.01344},
  url          = {https://arxiv.org/abs/2202.01344}
}

@misc{proofnet,
  title        = {ProofNet: Autoformalizing and Formally Proving Undergraduate-Level Mathematics},
  author       = {Azerbayev, Zhangir and Piotrowski, Bartosz and Schoelkopf, Hailey and Ayers, Edward W. and Radev, Dragomir and Avigad, Jeremy},
  year         = {2023},
  eprint       = {2302.12433},
  archivePrefix= {arXiv},
  primaryClass = {cs.CL},
  doi          = {10.48550/arXiv.2302.12433},
  url          = {https://arxiv.org/abs/2302.12433}
}

@misc{deepseekprover,
  title        = {DeepSeek-Prover: Advancing Theorem Proving in LLMs through Large-Scale Synthetic Data},
  author       = {Xin, Huajian and Guo, Daya and Shao, Zhihong and Ren, Z. Z. and Zhu, Qihao and Liu, Bo and Ruan, Chong and Li, Wenda and Liang, Xiaodan},
  year         = {2024},
  eprint       = {2405.14333},
  archivePrefix= {arXiv},
  primaryClass = {cs.CL},
  doi          = {10.48550/arXiv.2405.14333},
  url          = {https://arxiv.org/abs/2405.14333}
}

@misc{deepseekproverv2,
  title        = {DeepSeek-Prover-V2: Advancing Formal Mathematical Reasoning via Reinforcement Learning for Subgoal Decomposition},
  author       = {Ren, Z. Z. and Shao, Zhihong and Song, Junxiao and Xin, Huajian and Wang, Haocheng and Zhao, Wanjia and Zhang, Liyue and Fu, Zhe and Zhu, Qihao and Yang, Dejian and Wu, Z. F. and Gou, Zhibin and Ma, Shirong and Tang, Hongxuan and Liu, Yuxuan and Gao, Wenjun and Guo, Daya and Ruan, Chong},
  year         = {2025},
  eprint       = {2504.21801},
  archivePrefix= {arXiv},
  primaryClass = {cs.CL},
  doi          = {10.48550/arXiv.2504.21801},
  url          = {https://arxiv.org/abs/2504.21801}
}

@article{alphageometry,
  title   = {Solving olympiad geometry without human demonstrations},
  author  = {Trinh, Trieu H. and Wu, Yuhuai and Le, Quoc V. and He, He and Luong, Thang},
  journal = {Nature},
  year    = {2024},
  volume  = {625},
  number  = {7995},
  pages   = {476--482},
  doi     = {10.1038/s41586-023-06747-5},
  url     = {https://www.nature.com/articles/s41586-023-06747-5}
}

@misc{alphageometry2,
  title        = {Gold-medalist Performance in Solving Olympiad Geometry with AlphaGeometry2},
  author       = {Chervonyi, Yuri and Trinh, Trieu H. and Ol{\v{s}}{\'a}k, Miroslav and Yang, Xiaomeng and Nguyen, Hoang and Menegali, Marcelo and Jung, Junehyuk and Verma, Vikas and Le, Quoc V. and Luong, Thang},
  year         = {2025},
  eprint       = {2502.03544},
  archivePrefix= {arXiv},
  primaryClass = {cs.AI},
  doi          = {10.48550/arXiv.2502.03544},
  url          = {https://arxiv.org/abs/2502.03544}
}

@inproceedings{lean4,
  title     = {The Lean 4 Theorem Prover and Programming Language (System Description)},
  author    = {de Moura, Leonardo and Ullrich, Sebastian},
  booktitle = {International Conference on Automated Deduction (CADE)},
  year      = {2021},
  doi       = {10.1007/978-3-030-79876-5_37},
  url       = {https://lean-lang.org/papers/lean4.pdf}
}

@manual{coqmanual,
  title        = {The Coq Proof Assistant: Reference Manual},
  author       = {{The Coq Development Team}},
  year         = {2013},
  organization = {INRIA / TypiCal Project},
  note         = {Version 8.4pl2, April 4, 2013},
  url          = {https://flint.cs.yale.edu/cs430/coq/pdf/Reference-Manual.pdf}
}

@inproceedings{neculaPCC,
  title     = {Proof-Carrying Code},
  author    = {Necula, George C.},
  booktitle = {Proceedings of the 24th ACM SIGPLAN-SIGACT Symposium on Principles of Programming Languages (POPL)},
  year      = {1997},
  doi       = {10.1145/263699.263712},
  url       = {https://dl.acm.org/doi/10.1145/263699.263712}
}

@book{gordon1979lcf,
  title     = {Edinburgh LCF: A Mechanized Logic of Computation},
  author    = {Gordon, Michael J. C. and Milner, Arthur J. and Wadsworth, Christopher P.},
  series    = {Lecture Notes in Computer Science},
  volume    = {78},
  year      = {1979},
  publisher = {Springer},
  doi       = {10.1007/3-540-09724-4},
  url       = {https://link.springer.com/book/10.1007/3-540-09724-4}
}

@inproceedings{sledgehammer2010,
  title     = {Sledgehammer: Judgement Day},
  author    = {B{\"o}hme, Sascha and Nipkow, Tobias},
  booktitle = {Automated Reasoning (IJCAR 2010)},
  series    = {Lecture Notes in Computer Science},
  volume    = {6173},
  pages     = {107--121},
  year      = {2010},
  publisher = {Springer},
  doi       = {10.1007/978-3-642-14203-1_9},
  url       = {https://link.springer.com/chapter/10.1007/978-3-642-14203-1_9}
}

@article{sledgehammerSMT2013,
  title   = {Extending Sledgehammer with SMT Solvers},
  author  = {Blanchette, Jasmin Christian and B{\"o}hme, Sascha and Paulson, Lawrence C.},
  journal = {Journal of Automated Reasoning},
  volume  = {51},
  number  = {1},
  pages   = {109--128},
  year    = {2013},
  doi     = {10.1007/s10817-013-9278-5},
  url     = {https://link.springer.com/article/10.1007/s10817-013-9278-5}
}

@inproceedings{smtcoq2017,
  title     = {SMTCoq: A Plug-In for Integrating SMT Solvers into Coq},
  author    = {Ekici, Burak and Mebsout, Alain and Tinelli, Cesare and Keller, Chantal and Katz, Guy and Reynolds, Andrew and Barrett, Clark},
  booktitle = {Computer Aided Verification (CAV 2017)},
  series    = {Lecture Notes in Computer Science},
  year      = {2017},
  publisher = {Springer},
  doi       = {10.1007/978-3-319-63390-9_7},
  url       = {https://link.springer.com/chapter/10.1007/978-3-319-63390-9_7}
}

@inproceedings{holstep2017,
  title     = {HolStep: A Machine Learning Dataset for Higher-order Logic Theorem Proving},
  author    = {Kaliszyk, Cezary and Chollet, Fran{\c{c}}ois and Szegedy, Christian},
  booktitle = {International Conference on Learning Representations (ICLR)},
  year      = {2017},
  url       = {https://openreview.net/forum?id=ryuxYmvel}
}

@article{tactictoe2021,
  title   = {TacticToe: Learning to Prove with Tactics},
  author  = {Gauthier, Thibault and Kaliszyk, Cezary and Urban, Josef and Kumar, Ramana and Norrish, Michael},
  journal = {Journal of Automated Reasoning},
  volume  = {65},
  pages   = {257--286},
  year    = {2021},
  doi     = {10.1007/s10817-020-09580-x},
  url     = {https://dl.acm.org/doi/10.1007/s10817-020-09580-x}
}

@inproceedings{holist2019,
  title     = {{HOL}ist: An Environment for Machine Learning of Higher Order Logic Theorem Proving},
  author    = {Bansal, Kshitij and Loos, Sarah and Rabe, Markus and Szegedy, Christian and Wilcox, Stewart},
  booktitle = {Proceedings of the 36th International Conference on Machine Learning (ICML)},
  series    = {Proceedings of Machine Learning Research},
  volume    = {97},
  pages     = {454--463},
  year      = {2019},
  publisher = {PMLR},
  url       = {https://proceedings.mlr.press/v97/bansal19a.html}
}

@inproceedings{coqgym2019,
  title     = {Learning to Prove Theorems via Interacting with Proof Assistants},
  author    = {Yang, Kaiyu and Deng, Jia},
  booktitle = {Proceedings of the 36th International Conference on Machine Learning (ICML)},
  series    = {Proceedings of Machine Learning Research},
  volume    = {97},
  pages     = {6984--6994},
  year      = {2019},
  publisher = {PMLR},
  url       = {https://proceedings.mlr.press/v97/yang19a.html}
}

@inproceedings{proverbot90012020,
  title     = {Generating Correctness Proofs with Neural Networks},
  author    = {Sanchez-Stern, Alex and Alhessi, Yousef and Saul, Lawrence and Lerner, Sorin},
  booktitle = {Proceedings of the 4th ACM SIGPLAN International Workshop on Machine Learning and Programming Languages (MAPL)},
  year      = {2020},
  doi       = {10.1145/3394450.3397466},
  url       = {https://dl.acm.org/doi/10.1145/3394450.3397466}
}

@inproceedings{tacticzero2021,
  title     = {TacticZero: Learning to Prove Theorems from Scratch with Deep Reinforcement Learning},
  author    = {Wu, Minchao and Norrish, Michael and Walder, Christian and Dezfouli, Amir},
  booktitle = {Advances in Neural Information Processing Systems (NeurIPS)},
  year      = {2021},
  url       = {https://proceedings.neurips.cc/paper/2021/hash/4dea382d82666332fb564f2e711cbc71-Abstract.html}
}

@misc{gptf2020,
  title         = {Generative Language Modeling for Automated Theorem Proving},
  author        = {Polu, Stanislas and Sutskever, Ilya},
  year          = {2020},
  eprint        = {2009.03393},
  archivePrefix = {arXiv},
  primaryClass  = {cs.LG},
  doi           = {10.48550/arXiv.2009.03393},
  url           = {https://arxiv.org/abs/2009.03393}
}

@inproceedings{minif2f2022,
  title         = {MiniF2F: a Cross-System Benchmark for Formal Olympiad-Level Mathematics},
  author        = {Zheng, Kunhao and Han, Jesse Michael and Polu, Stanislas},
  booktitle     = {International Conference on Learning Representations (ICLR)},
  year          = {2022},
  eprint        = {2109.00110},
  archivePrefix = {arXiv},
  doi           = {10.48550/arXiv.2109.00110},
  url           = {https://arxiv.org/abs/2109.00110}
}

@misc{formalStmtCL2022,
  title         = {Formal Mathematics Statement Curriculum Learning},
  author        = {Polu, Stanislas and Han, Jesse Michael and Zheng, Kunhao and Baksys, Mantas and Babuschkin, Igor and Sutskever, Ilya},
  year          = {2022},
  eprint        = {2202.01344},
  archivePrefix = {arXiv},
  primaryClass  = {cs.LG},
  doi           = {10.48550/arXiv.2202.01344},
  url           = {https://arxiv.org/abs/2202.01344}
}

@misc{leandojo2023,
  title         = {LeanDojo: Theorem Proving with Retrieval-Augmented Language Models},
  author        = {Yang, Kaiyu and Swope, Aidan M. and Gu, Alex and Chalamala, Rahul and Song, Peiyang and Yu, Shixing and Godil, Saad and Prenger, Ryan and Anandkumar, Anima},
  year          = {2023},
  eprint        = {2306.15626},
  archivePrefix = {arXiv},
  primaryClass  = {cs.LG},
  doi           = {10.48550/arXiv.2306.15626},
  url           = {https://arxiv.org/abs/2306.15626}
}

@misc{deepseekprover2024,
  title         = {DeepSeek-Prover: Advancing Theorem Proving in LLMs through Large-Scale Synthetic Data},
  author        = {Xin, Huajian and Guo, Daya and Shao, Zhihong and Ren, Zhizhou and Zhu, Qihao and Liu, Bo and Ruan, Chong and Li, Wenda and Liang, Xiaodan},
  year          = {2024},
  eprint        = {2405.14333},
  archivePrefix = {arXiv},
  primaryClass  = {cs.CL},
  doi           = {10.48550/arXiv.2405.14333},
  url           = {https://arxiv.org/abs/2405.14333}
}

@misc{deepseekproverv22025,
  title         = {DeepSeek-Prover-V2: Advancing Formal Mathematical Reasoning via Reinforcement Learning for Subgoal Decomposition},
  author        = {Ren, Z. Z. and Shao, Zhihong and Song, Junxiao and Xin, Huajian and Wang, Haocheng and Zhao, Wanjia and Zhang, Liyue and Fu, Zhe and Zhu, Qihao and Yang, Dejian and Wu, Z. F. and Gou, Zhibin and Ma, Shirong and Tang, Hongxuan and Liu, Yuxuan and Gao, Wenjun and Guo, Daya and Ruan, Chong},
  year          = {2025},
  eprint        = {2504.21801},
  archivePrefix = {arXiv},
  primaryClass  = {cs.AI},
  doi           = {10.48550/arXiv.2504.21801},
  url           = {https://arxiv.org/abs/2504.21801}
}

@misc{proofnet2023,
  title         = {ProofNet: Autoformalizing and Formally Proving Undergraduate-Level Mathematics},
  author        = {Azerbayev, Zhangir and Piotrowski, Bartosz and Schoelkopf, Hailey and Ayers, Edward W. and Radev, Dragomir and Avigad, Jeremy},
  year          = {2023},
  eprint        = {2302.12433},
  archivePrefix = {arXiv},
  primaryClass  = {cs.CL},
  doi           = {10.48550/arXiv.2302.12433},
  url           = {https://arxiv.org/abs/2302.12433}
}

@article{alphageometry2024,
  title   = {Solving olympiad geometry without human demonstrations},
  author  = {Trinh, Trieu H. and Wu, Yuhuai and Le, Quoc V. and Luong, Thang},
  journal = {Nature},
  year    = {2024},
  volume  = {625},
  number  = {7995},
  pages   = {476--482},
  doi     = {10.1038/s41586-023-06747-5},
  url     = {https://www.nature.com/articles/s41586-023-06747-5}
}

@misc{alphageometry22025,
  title         = {Gold-medalist Performance in Solving Olympiad Geometry with AlphaGeometry2},
  author        = {Chervonyi, Yuri and Trinh, Trieu H. and Ol{\v{s}}{\'a}k, Miroslav and Yang, Xiaomeng and Nguyen, Hoang and Menegali, Marcelo and Jung, Junehyuk and Verma, Vikas and Le, Quoc V. and Luong, Thang},
  year          = {2025},
  eprint        = {2502.03544},
  archivePrefix = {arXiv},
  primaryClass  = {cs.AI},
  doi           = {10.48550/arXiv.2502.03544},
  url           = {https://arxiv.org/abs/2502.03544}
}

@inproceedings{neculaPCC1997,
  title     = {Proof-Carrying Code},
  author    = {Necula, George C.},
  booktitle = {Proceedings of the 24th ACM SIGPLAN-SIGACT Symposium on Principles of Programming Languages (POPL)},
  year      = {1997},
  doi       = {10.1145/263699.263712},
  url       = {https://dl.acm.org/doi/10.1145/263699.263712}
}

@article{lfsc2013,
  title   = {SMT Proof Checking Using a Logical Framework},
  author  = {Stump, Aaron and Oe, Duckki and Reynolds, Andrew and Hadarean, Liana and Tinelli, Cesare},
  journal = {Formal Methods in System Design},
  volume  = {42},
  number  = {1},
  pages   = {91--118},
  year    = {2013},
  doi     = {10.1007/s10703-012-0163-3},
  url     = {https://dl.acm.org/doi/10.1007/s10703-012-0163-3}
}

@inproceedings{drattrim2014,
  title     = {DRAT-trim: Efficient Checking and Trimming Using Expressive Clausal Proofs},
  author    = {Wetzler, Nathan and Heule, Marijn J. H. and Hunt, Warren A.},
  booktitle = {Theory and Applications of Satisfiability Testing -- SAT 2014},
  series    = {Lecture Notes in Computer Science},
  volume    = {8561},
  pages     = {422--429},
  year      = {2014},
  publisher = {Springer},
  doi       = {10.1007/978-3-319-09284-3_31},
  url       = {https://link.springer.com/chapter/10.1007/978-3-319-09284-3_31}
}

\end{document}